%% file: main.tex
\documentclass[sigconf,nonacm]{acmart}
\AtBeginDocument{%
  }

\usepackage{subcaption}
\begin{document}

%%
%% The "title" command has an optional parameter,
%% allowing the author to define a "short title" to be used in page headers.
\title{Long-term User Engagement Optimization through Model-agnostic Downstream Rewards Learning}

%%
%% The "author" command and its associated commands are used to define
%% the authors and their affiliations.
%% Of note is the shared affiliation of the first two authors, and the
%% "authornote" and "authornotemark" commands
%% used to denote shared contribution to the research.

\author{Dingsu Wang}
\authornote{All authors contributed equally to this research work.}
\email{derekwang@pinterest.com}
\author{Filip Ryzner}
\authornotemark[1]
\email{fryzner@pinterest.com}
\author{Kelly He}
\authornotemark[1]
\email{kellyhe@pinterest.com}
\author{Armando Ordorica}
\email{aordorica@pinterest.com}
\authornotemark[1]
\author{David Woo}
\email{dwoo@pinterest.com}
\authornote{Corresponding author.}
\author{Aditya Mantha}
\email{amantha@pinterest.com}
\author{Liyao Lu}
\email{llu@pinterest.com}
\affiliation{
  \institution{Pinterest}
  \city{San Francisco}
  \state{CA}
  \country{USA}
}
\author{Usha Amrutha Nookala}
\email{unookala@pinterest.com}
\author{Haoran Guo}
\email{haoranguo@pinterest.com}
\authornote{Work done at Pinterest.}
\author{Jiacong He}
\email{jiaconghe@pinterest.com}
\authornotemark[3]
\author{Olafur Gudmundsson}
\email{ogudmundsson@pinterest.com}
\author{Matt Chun}
\email{mchun@pinterest.com}
\author{Krystal Benitez}
\email{kbenitez@pinterest.com}
\author{Haibin Xie}
\email{hxie@pinterest.com}
\affiliation{
  \institution{Pinterest}
  \city{San Francisco}
  \state{CA}
  \country{USA}
}

\author{Alekhya Pyla}
\email{apyla@pinterest.com}
\author{Sameer Jain}
\email{sameerjain@pinterest.com}
\author{Zhongjian Jiang}
\email{zhongjianjiang@pinterest.com}
\author{Shruthi Hariharan}
\email{shariharan@pinterest.com}
\author{Dhruvil Deven Badani}
\email{dbadani@pinterest.com}
\author{Yijie Dylan Wang}
\email{dylanwang@pinterest.com}
\affiliation{
  \institution{Pinterest}
  \city{San Francisco}
  \state{CA}
  \country{USA}
}

% \author{Jaewon Yang}
% \email{jaewonyang@pinterest.com}
% \affiliation{
%   \institution{Pinterest}
%   \city{San Francisco}
%   \state{CA}
%   \country{USA}
% }

%%
%% By default, the full list of authors will be used in the page
%% headers. Often, this list is too long, and will overlap
%% other information printed in the page headers. This command allows
%% the author to define a more concise list
%% of authors' names for this purpose.
\renewcommand{\shortauthors}{Dingsu Wang et al.}

%%
%% The code below is generated by the tool at http://dl.acm.org/ccs.cfm.
%% Please copy and paste the code instead of the example below.
%%


%%
%% Keywords. The author(s) should pick words that accurately describe
%% the work being presented. Separate the keywords with commas.

% \received{20 February 2007}
% \received[revised]{12 March 2009}
% \received[accepted]{5 June 2009}

%%
%% This command processes the author and affiliation and title
%% information and builds the first part of the formatted document.

\input{sections/0_abstract}

\maketitle
\let\thefootnote\relax\footnotetext{Note: This manuscript is a revised version of the paper accepted for publication at RecSys 2026.}

\input{sections/1_intro}

\input{sections/3_preliminaries}

\input{sections/4_analysis}
\input{sections/5_methods}

\input{sections/6_experiment}
\input{sections/2_related}

\input{sections/7_conclusions}
\input{sections/8_acknowledgement}

%%
%% The next two lines define the bibliography style to be used, and
%% the bibliography file.
\bibliographystyle{ACM-Reference-Format}
\bibliography{references}

\input{sections/9_appendix}
\end{document}

%% file: sections/0_abstract.tex
\begin{abstract}
As recommender systems mature in the past few years, their optimization objectives have evolved from a primary focusing on short-term behavioral signals to a broader emphasis on long-term user engagement and retention. However, directly optimizing retention is difficult because return signals are sparse, delayed, and only partially attributable to earlier recommendations. Prior work has addressed this challenge with sequential modeling and reinforcement learning, but these approaches typically require task specific reward engineering, substantial computational overhead, and surface specific implementations that are difficult to generalize. In this paper, we present a unified, model-agnostic downstream reward framework for optimizing long-term user value in large-scale recommendation systems. First, we formulate the downstream reward learning problem and develop an offline screening framework to identify session level behaviors that are both observable early and predictive of future retention. We then propose several model-agnostic downstream rewards signals derived from observed user action patterns across multiple sources. We further discuss the engineering effort to productionize the proposed rewards derivations and challenges we faced when adding them to our ranking models. Online A/B experiments demonstrate consistent improvements in engagement and retention-related metrics, and the framework has been deployed across multiple Pinterest surfaces, including Homefeed, Related Pins, Search, and Notifications.
\end{abstract}

%% file: sections/1_intro.tex
\section{Introduction}

Many modern recommendation systems have proven to be effective at optimizing short-term user actions. However, optimizing for immediate engagement alone could potentially hurt long-term user value. Prior works argue that feedback loops driven by short-term interaction signals may increase homogeneity and hurt user return behavior over time \cite{sculley2015hidden,wu2017returning,chaney2018algorithmic}. For large-scale recommendation platforms such as Pinterest, the core challenge is therefore not only to predict what a user will do next immediately, but also to optimize recommendations for longer-term engagement and retention.

Direct optimization of long-term retention is difficult for several reasons. First of all, retention labels are sparse, noisy, and often only weakly observable \cite{wang2022surrogate}. Secondly, retention is delayed: even when downstream outcomes are observed, attributing them to specific earlier recommendations is challenging \cite{cai2023reinforcing,xue2023prefrec}. Finally, user interests and data distributions evolve over time, so fixed surrogate objectives can become stale and lose alignment with the behaviors that actually drive long-term value \cite{yoo2025generalizable}.

Recent work has approached this problem using reinforcement learning (RL) and related long-horizon optimization methods \cite{zhang2021counterfactual,wang2022surrogate,zhang2022multi,cai2023reinforcing,xu2023optimizing,xue2023prefrec,xue2025auro}. These methods are appealing because they explicitly optimize accumulated future reward. However, deploying RL in production recommendation systems remains difficult: it requires large amounts of interaction data \cite{chen2024opportunities}, must operate over extremely large action spaces \cite{christakopoulou2022reward}, and depends on reward designs that may not transfer cleanly across surfaces with different product goals, such as Homefeed, Search, and Notifications at Pinterest.

\begin{figure}[t]
    \centering
    \includegraphics[width=0.95\linewidth]{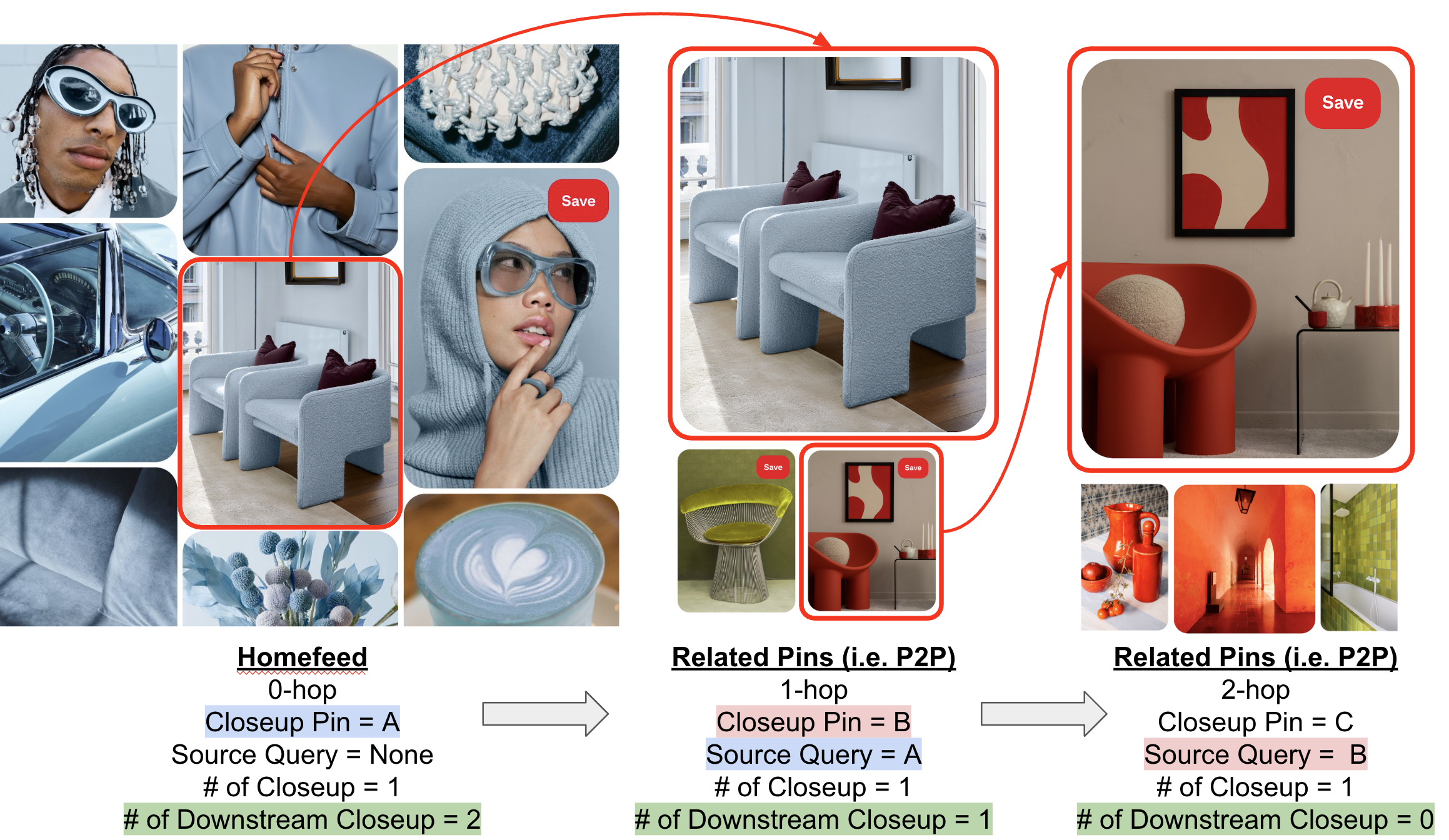}
    \caption{An illustration of user action patterns in P2P rabbit hole with closeup action at Pinterest.}
    \label{fig:p2p_nhop_closeup}
    \Description[P2P N-Hop Closeup actions]{User closeup actions in P2P rabbit hole at Pinterest.}
\end{figure}

To tackle all the challenges above, we propose a unified model-agnostic solution based on downstream rewards (DR) modeling. Our method have two benefits: (1) it could work with any RL or supervised-learning based recommendation models across different surfaces, (2) the proxy rewards we proposed can help drive long-term retention optimization. Specifically, we study billions of user activity data from Pinterest, and we propose to use downstream rewards signals from P2P (i.e. Pin to Pin) rabbit hole, negative user engagements and user interests from engagement sequences as an proxy optimization signal to maximize the long-term user engagements and retention. An illustration of user actions in P2P rabbit hole could be found in Figure~\ref{fig:p2p_nhop_closeup}. Our approach is motivated by offline analysis showing that deeper P2P exploration, saves, and deep engagement across diverse content are strongly associated with future engagements and retention, while shallow high-volume browsing is often misleading and noisy. These downstream rewards are observed earlier and at much higher volume than long-term retention labels, which makes them more practical for large-scale learning. The proposed downstream rewards consist of both positive and negative user action signals which cover majority of important action patterns at Pinterest. Since these semi-orthogonal downstream rewards are observed along the user’s near-term interactions sequences, they provide denser and less delayed information than long-term retention labels, which helps alleviate the label sparsity and attribution challenges. Moreover, since the downstream rewards are derived from observed engagements, they remain aligned with evolving feature distributions and user behaviors over time which help model capture this change and mitigate the data distribution shift challenge. Finally, we also discuss the infrastructure optimization for deriving the labels at scale efficiently in the paper. To validate the effectiveness of the proposed DR signals, we experiment with incorporating them in a production recommendation system at Pinterest \cite{xia2023transact}. Results from long-time online A/B testing experiment show that our methods significantly improve the user engagement and retention across different platforms.

The main contributions of this paper can be summarized as:
\begin{itemize}
    \item \textbf{Insights for User Retention Modeling.} We present an offline evaluation framework that identifies \emph{which} session-level signals should serve as downstream reward. The framework screens candidates against three criteria (indicative of long-term retention, observable at the session level, and incrementally predictive in a multivariate model) and narrows the search space from hundreds of candidates to a smaller set for online experimentation.
    \item \textbf{Model-agnostic Downstream Rewards.} We propose a unified, model-agnostic downstream reward framework for optimizing long-term user engagement and retention that can be applied to different deep learning based recommendation models across surfaces. The framework combines complementary rewards that capture multiple aspects of user value and transfer readily across surfaces.
    \item \textbf{Reward Label Generation.} We built an efficient infrastructure to generate the downstream rewards signal based on multiple level of engagement sequences so that recommendation models from different platforms could easily adopt.
    \item \textbf{Empirical Evaluations.} We conducted online A/B testing on a real-world recommendation system using the proposed DR modeling methods to demonstrate its effectiveness.
\end{itemize}
The rest of the paper is organized as follows: Problem definitions are included in Section~\ref{sec:preliminary} and data insights for our downstream rewards as well as analysis are listed in Section~\ref{sec:insights}. Section~\ref{sec:methodology} presents the proposed downstream reward designs as well as engineering effort to generate them in productions. Experiments and the reviews of related works are given in Section~\ref{sec:experiments} and Section~\ref{sec:related} respectively. Section~\ref{sec:conclusion} concludes the paper.

%% file: sections/3_preliminaries.tex
\section{Problem Definition}\label{sec:preliminary}

\begin{definition}[\textbf{Optimal User Retention Objective}]\label{def:optimal}
    Let $\textbf{H}_{u,t}$ denote the available historical data for user $u$ and its engaged Pins before time $t$, and $x_{u,t} \sim \pi_\theta(\cdot | \mathbf{H}_{u,t})$ denote the candidate recommended Pin for user $u$ at time t by the recommendation policy $\pi_\theta$. $\mathcal{R}_{u,t} \in \{ 0, 1 \}$ denote the conditional reward: whether user $u$ will return and remain active or take any action within time window $\Delta$ after time $t$. Its conditional distribution is induced by user dynamics given past history $\mathbf{H}_{u,t}$ and recommended candidate Pin $x_{u,t}$. The ideal objective for maximizing user $u$'s retention can be defined as:
    \begin{equation*}
        \mathcal{J}_{u}(\theta) =  \mathbb{E}_{\tau \sim (\pi_\theta, P_u)} \left[\sum_{t=0}^{T} \gamma^{t} \mathcal{R}_{u,t}(\mathbf{H}_{u,t}, x_{u,t})\right],
    \end{equation*}
    where $\gamma \in (0, 1]$ is the discount factor, and $P_u$ is the user dynamics. $\tau$ here stands for the interaction trajectory inferred by  the recommendation model and the user corresponding engagements.
\end{definition}

Based on the above definition, then the goal is to find optimal recommendation models $\theta^\star$ that $\theta^\star \in \arg\max_{\theta} \mathcal{J}_u(\theta)$

\begin{definition}[\textbf{Proxy Downstream Rewards}]\label{def:proxy-p2p}
    Let $\textbf{S}_{u,t}$ denote the user $u$'s action trajectory in the P2P rabbit hole after closeup on the source Pin $x_{u,t}$ at time $t$, which can be defined as:
    \begin{equation*}
        \textbf{S}_{u,t} = \left((x_{u,t}, \texttt{closeup}), (x_{u,t+i}, A_{u,t+i}), ..., (x_{u,t+n}, A_{u,t+n}) \right)
    \end{equation*}
    where $A_{u,t+i}$ represents user $u$'s action on Pin $x_{u,t+i}$ which could be $\texttt{click}$, $\texttt{closeup}$, $\texttt{save}$, $\texttt{download}$ etc. Let the proxy learned P2P entry model be $p_{\beta}^{enter}(\mathcal{E}|\textbf{H}_{u,t}, x_{u,t})$ where $\mathcal{E} \in \{0, 1 \}$ indicates whether the user will closeup on $x_{u,t}$, and let $p_\phi(\mathbf{S}_{u,t} | \mathbf{H}_{u,t}, x_{u,t}, \mathcal{E} {=} 1)$ be the learned proxy downstream trajectory prediction model. Here, $\gamma$ and $\phi$ denote the learnable parameter sets of the entry model and the downstream trajectory model respectively. Then, the \textit{proxy} downstream rewards is:
    \begin{equation*}
    \begin{split}
        \hat{\mathcal{R}}_{u,t}(\mathbf{H}_{u,t}, x_{u,t}) = \mathbb{E}_{\mathcal{E}\sim p_{\beta}^{enter}} \Big[ \mathbb{E}_{\hat{\mathbf{S}}\sim p_\phi} \big[\mathbb{R}(\hat{\mathbf{S}})\big] \Big]
    \end{split}
    \end{equation*}
    where $\mathbb{R}$ is a surrogate rewards aggregation function that depends on actions in the predicted trajectory $\hat{\mathbf{S}}$ such as sum of action rewards $\mathbb{R}(\hat{\mathbf{S}}) = \sum_{i=1}^{n} r\!\left(\hat{A}_{u,t+i}\right)$ where $r(\texttt{closeup})=r(\texttt{exit})=0$. $\mathbb{R}(\hat{\mathbf{S}})=0$ if $\mathcal{E}=0$, which means user never enters P2P rabbit hole.
\end{definition}

While Definition~\ref{def:proxy-p2p} is generic, its implementation varies across surfaces. On Homefeed and Search, ranking is applied to source candidate Pins $x_{u,t}$, so the optimized predictions are primarily tied to the entry model $p_{\beta}^{\mathrm{enter}}$, which estimates whether a candidate will trigger closeup and lead to a valuable downstream P2P trajectory. On Related Pins, ranking occurs within the P2P rabbit hole itself, so the implementation is more closely aligned with $p_{\phi}$, which models the quality of the subsequent sequence of related Pins interactions. Other downstream rewards in this paper, such as negative rewards and rewards for use-case adoption, can be defined within the same framework by modifying the action sequence $\mathbf{S}_{u,t}$, the reward function $\mathbb{R}$, and the time horizon $T$.

\begin{definition}[\textbf{Proxy User Retention Objective}]
    Similar to Definition \ref{def:optimal}, the proxy user retention objective based on downstream rewards can be defined as:
    \begin{equation*}
        \hat{\mathcal{J}}_{u}(\theta;\beta,\phi) =  \mathbb{E}_{\tau \sim (\pi_\theta, P_u)} \left[\sum_{t=0}^{T} \beta^{t} \hat{\mathcal{R}}_{u,t}(\mathbf{H}_{u,t}, x_{u,t})\right]
    \end{equation*}
    % \david{wondering if we need the proxy user retention obj or the optimal user retention objective would be enough?}
\end{definition}

%% file: sections/4_analysis.tex
\section{Insights and Analysis}\label{sec:insights}

% \subsection{Offline Identification of Retention-Aligned Reward Proxies}\label{sec:analysis}

Directly optimizing long-term retention is difficult because revisit labels are sparse, delayed, and only partially attributable to any recommendation. We therefore use an offline screening framework to identify \emph{session-level} behaviors that are sufficiently aligned with future revisitation to serve as proxy downstream rewards. A candidate reward should (1) correlate with future revisitation rather than only immediate activity, (2) be observable within or shortly after the session, and (3) remain predictive after controlling for correlated signals. The output is a ranked list of candidates for online testing.

\paragraph{Data and target.}
Each training and evaluation example is a per-user daily observation centered on a "pivot day" $t=0$. We focus on users in a stable low-engagement state before the pivot. Let $D_u(i)$ be the number of active days for user $u$ in week $i$ relative to the pivot. We define a user to have low engagement when $D_u(-2)\leq 1$ and $D_u(-1)\leq 1$ , and label a positive transition by
$Y_u=1 \hspace{0.5em} \text{if} \hspace{0.5em} D_u(+1)\geq 3 \ \text{and}\ D_u(+2)\geq 3.$ About $4\%$ of the total low-engagement population satisfies this criteria. The task for our offline analysis is to predict whether the pivot day behavior precedes a transition from low engagement to sustained higher engagement.

\paragraph{Features and normalization.}
For each "pivot day", we extract cross-session, cross-surface activity features, including impressions, closeups, saves, clicks, search effort, and diversity signals, expanding 174 base features into $\sim 10^3$ crossed candidates. Because raw counts depend strongly on exposure, we normalize each feature by comparing its pivot-day action density to the user’s recent baseline: $r_{u,0}^{(j)} = a_{u,0}^{(j)} / I_{u,0}$ and $\hat{x}_{u,0}^{(j)} = r_{u,0}^{(j)} - \tilde{r}_u^{(j)}$, where $\tilde{r}_u^{(j)}$ is the user-specific median action density over the prior two weeks. We call this \emph{user-prevalence} (UP) normalization; $\hat{x}_{u,0}^{(j)} > 0$ and $\hat{x}_{u,0}^{(j)} < 0$ indicate above and below baseline engagement respectively. We also impute missing values and remove highly collinear features. For comparison in Table~\ref{tab:feature-summary}, we report results with \emph{percentile normalization} (Pctl), which normalizes each feature to its percentile within the user’s historical distribution.

\paragraph{Screening model.}
We train a Random Forest on hundreds of thousands of pivot-day examples with held-out validation and test sets. We use the model for \emph{screening}: correlations, feature importances, SHAP values, and dependence analysis are used to rank candidate downstream rewards. On held-out data, the model achieves AUC $0.65$--$0.70$, which indicates sufficient predictive signal.

\begin{table}[ht]
\centering
\small
\begin{tabular}{@{}l cc cc c@{}}
\toprule
& \multicolumn{2}{c}{\textbf{Pctl.\ Norm.}} & \multicolumn{2}{c}{\textbf{UP Norm.}} & \textbf{RF} \\
\cmidrule(lr){2-3} \cmidrule(lr){4-5} \cmidrule(lr){6-6}
\textbf{Feature category} & \textbf{Sign} & \textbf{Str.} & \textbf{Sign} & \textbf{Str.} & \textbf{Imp.} \\
\midrule
Diverse (shallow)   & $+$ & H & $-$ & H & H \\
Diverse (deep)      & $+$ & H & $+$ & M & H \\
P2P engagement      & $+$ & H & $+$ & H & H \\
Pin clicks          & $+$ & H & $-$ & M & M \\
Saves              & $+$ & M & $+$ & H & M \\
P2P entry diversity & $+$ & M & $-$ & M & -- \\
Surface diversity   & $+$ & L & $-$ & L & M \\
\bottomrule
\end{tabular}
\caption{Summary of univariate correlations and Random Forest importance from different normalizations. Several shallow signals flip sign after controlling for impression volume, while P2P engagement, saves, and deep diverse engagement remain positive.}
\label{tab:feature-summary}
\end{table}

\paragraph{Main screening result.}
Table~\ref{tab:feature-summary} summarizes the main feature groups under two normalization schemes: \emph{Pctl.\ Norm.} and \emph{UP Norm.} as discussed above. For each scheme, \emph{Sign} indicates the direction of association with future sustained engagement, \emph{Str.} is its strength (high, medium, or low), and \emph{RF Imp.} is the feature-group importance from the Random Forest. A consistent pattern from results is that several shallow signals are positive under percentile normalization but negative under UP normalization, including shallow diverse engagement, Pin clicks, P2P entry diversity, and surface diversity. This suggests that they mainly capture broad but shallow browsing. In contrast, deep engagement remains positive under both schemes. P2P engagement, saves, and deep engagement over diverse content also rank highly in the Random Forest. Overall, \emph{how} users engage is more predictive than how much content they see.

Two findings are especially robust. First, \textbf{P2P exploration density correlates with faster return}: downstream rewards in P2P are a valid proxy for long-term user engagement. Second, \textbf{shallow and deep diversity have opposite effects}: lightly interacting with many categories is negative after exposure normalization, whereas deep exploration across different categories remains positive.

\begin{table}[ht]
\centering
\small
\begin{tabular}{@{}l c@{}}
\toprule
\textbf{Starting surface-action state} & \textbf{Expected feedviews} \\
\midrule
HF save+closeup      & 7.80 \\
RP save+closeup      & 7.45 \\
HF closeup            & 7.36 \\
Search save+closeup  & 7.26 \\
Board Feed closeup    & 6.45 \\
\midrule
Homefeed              & 4.45 \\
Related Pins          & 4.39 \\
Search                & 3.60 \\
\bottomrule
\end{tabular}
\caption{Representative expected downstream session lengths from the session-level Markov analysis. Deep states involving closeups and especially saves yield substantially longer continuation than surface-only entry states.}
\label{tab:session-length-states}
\end{table}

\paragraph{Deeper actions lead to longer sessions.}
The pivot-day model above operates on daily aggregates. To corroborate the same pattern at the session level, we analyze a Markov chain whose states are surface-action pairs, with session end as the terminal state, and estimate the expected number of subsequent feedviews\footnote{Each feedview is a request from user such as actions or multiple impressions etc.} transitions before termination from each starting state using the empirical transition matrix from production traffic. Table~\ref{tab:session-length-states} shows results of representative states. \emph{Deep} states dominate: closeup states extend sessions beyond surface only entries, and states that combine \texttt{save} and \texttt{closeup} rank highest. On Homefeed, \texttt{save}+\texttt{closeup} yields $7.80$ expected feedviews versus $4.45$ for Homefeed only.

\begin{figure}[ht]
  \centering
  \includegraphics[width=0.99\linewidth]{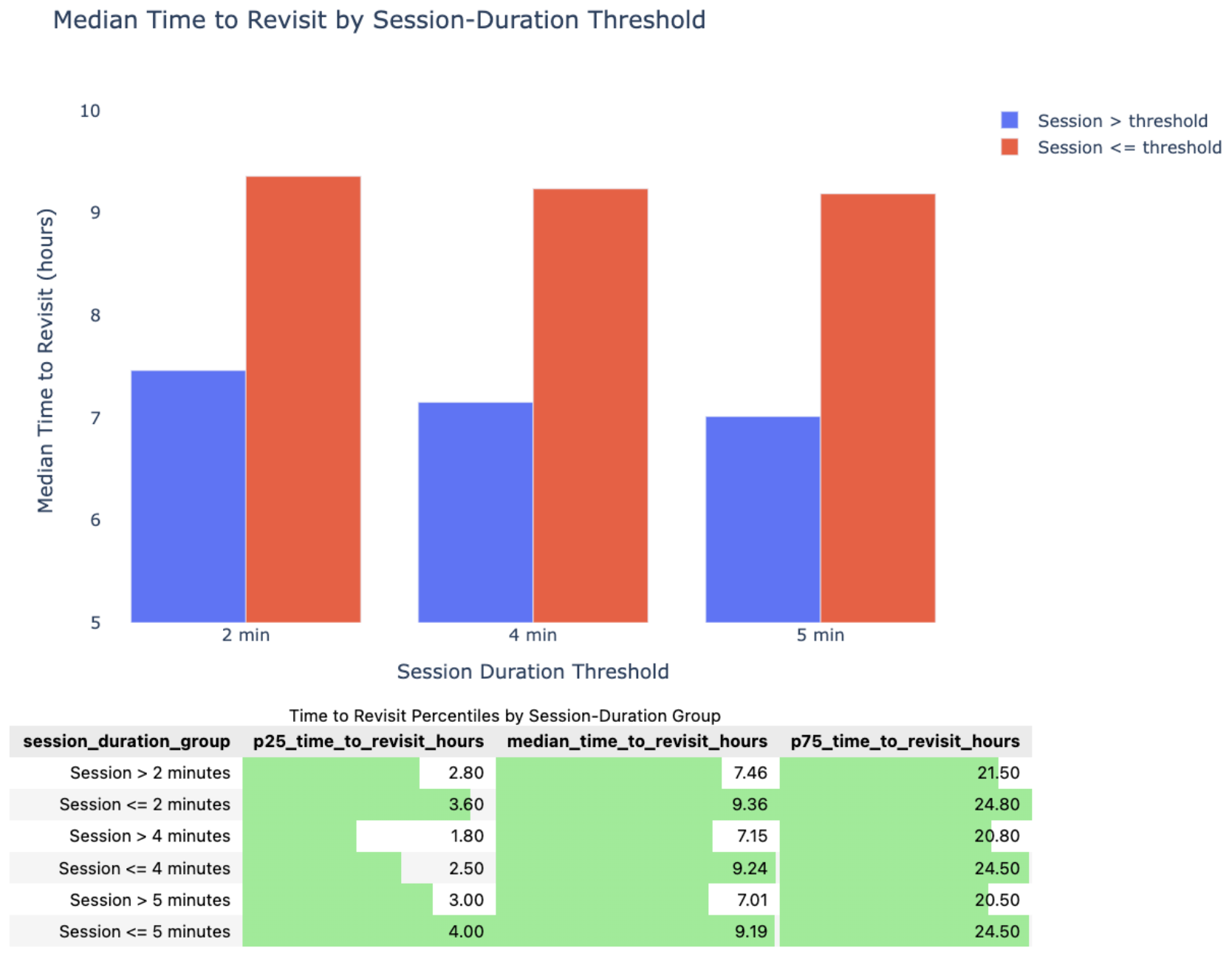}
  \caption{Median hours to next revisit for sessions above versus below each duration threshold. Longer sessions consistently revisit sooner.}
  \label{fig:time-to-revisit}
\end{figure}

\paragraph{Longer sessions lead to faster revisit.}
Finally, we test whether longer sessions are associated with faster return. Figure~\ref{fig:time-to-revisit} compares the median time to next revisit for sessions above and below a range of duration thresholds. Across all nine thresholds tested (2--38 minutes), longer sessions revisit significantly sooner (all $t$-tests satisfy $p < 10^{-43}$). The median gap is about $2.2$ hours at the 5-minute threshold and about $2.7$ hours at the 20-minute threshold. Although observational rather than causal, these results support a consistent empirical chain:
$$
\text{deeper actions} \Rightarrow \text{longer sessions} \Rightarrow \text{faster revisit} \Rightarrow \text{retention}\uparrow.
$$

\paragraph{Summary.}
The offline framework uses pivot-day behavior to predict later engagements pattern and then screens candidate reward metrics using the trained model. The main result is that deep P2P exploration, saves, and deep engagement across diverse content are the strongest positive proxies for long-term retention. In contrast, shallow high-volume browsing signals are often misleading once exposure is controlled for. These findings motivate downstream rewards that favor depth, cross-surface exploration, and revisitation-oriented actions rather than raw activity volume.

%% file: sections/5_methods.tex
\section{Methodology}\label{sec:methodology}
We approximate long-term user value with a set of complementary downstream rewards derived from near-term user behavior. As shown in Section~\ref{sec:insights}, the strongest proxies reflect multiple aspects of user experience, including deeper engagement and avoidance of low-quality interactions, while the value of exploration depends on whether it is shallow or deep. We therefore define three reward families: (1) deeper session engagement, (2) negative rewards, and (3) use case adoption. We then describe the infrastructures optimization efforts to derive these rewards at scale and the objective balance between DR signals and immediate action signals.

\subsection{Downstream Rewards}\label{sec:rewards}

\subsubsection{Rewards for Deeper Session Engagement}
In practice, a recommendation can contribute to user value not only by triggering a single high-intent action, but also by sustaining a series
of downstream interactions such as additional closeups, saves, long clicks, and P2P exploration. To capture this effect, we define a reward based on
the discounted cumulative engagement over the downstream session trajectory.

Specifically, consider a trajectory $\mathbf{S}_{u,t}$ as defined in
Definition~\ref{def:proxy-p2p}, consisting of downstream events after a recommendation at
time $t$:
$
\mathbf{S}_{u,t} = \left(e_{u,t+1}, e_{u,t+2}, \dots, e_{u,t+n_{u,t}}\right),
$
where $e_{u,t+i} = (x_{u,t+i}, A_{u,t+i})$ denotes the item-action pair at downstream step $i$, and $n_{u,t}$ is the number of downstream events until the end of the session. Let $\mathcal{A}^{\mathrm{eng}}$ denote the set of engagement actions of interest such as:
$
\mathcal{A}^{\mathrm{eng}} =
\{\texttt{save}, \texttt{downloads}, \texttt{screenshots}\}.
$
For each action $a \in \mathcal{A}^{\mathrm{eng}}$, let $w_a \ge 0$ denote its engagement value. We define the per-step engagement reward as
$
r_{u,t+i}^{\mathrm{eng}} = \sum_{a \in \mathcal{A}^{\mathrm{eng}}} w_a \, \mathbb{I}[A_{u,t+i} = a].
$

The deeper session engagement reward is then defined as the discounted sum of downstream
engagement:
\begin{equation*}
\mathbb{R}^{\mathrm{eng}}(\mathbf{S}_{u,t})
=
\sum_{i=1}^{n_{u,t}} \gamma^{i-1} r_{u,t+i}^{\mathrm{eng}},
\end{equation*}
where $\gamma \in (0,1]$ is a discount factor that assigns greater credit to engagement events that occur closer to the original recommended Pins, and $\mathbb{R}^{\mathrm{eng}}$ captures cumulative downstream engagement over the session. For a predicted trajectory $\hat{\mathbf{S}}$, we could analogously define $\mathbb{R}^{\mathrm{eng}}(\hat{\mathbf{S}})$ based on predicted per-step reward $\hat{r}_{u,t+i}^{\mathrm{eng}}$.

\subsubsection{Negative Rewards}\label{sec:negative-rewards}
\begin{figure}[ht]
\centering
\begin{subfigure}{0.98\columnwidth}
  \includegraphics[width=1\linewidth]{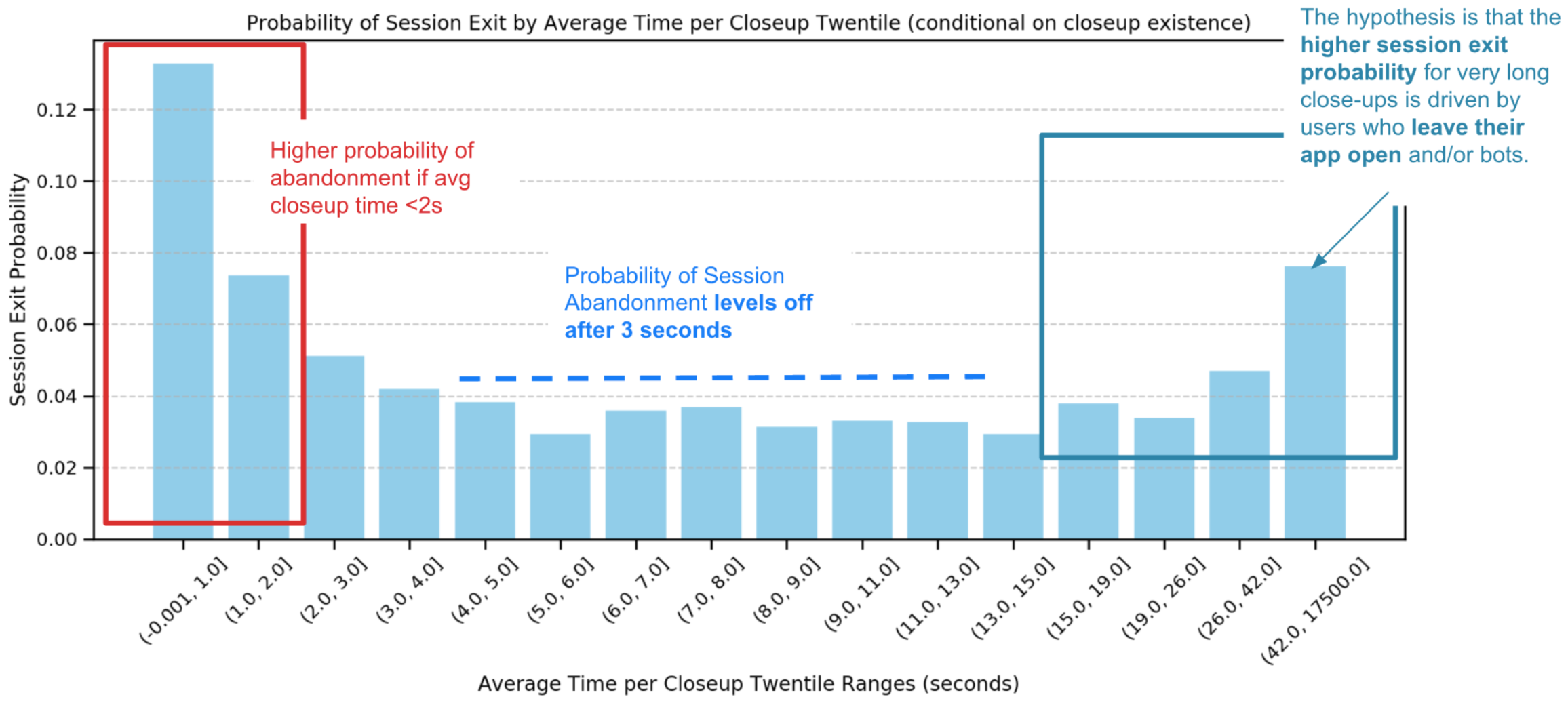}
  \caption{Probability of session exits of different closeup durations.}
\end{subfigure}
\medskip
\begin{subfigure}{0.98\columnwidth}
  % \includegraphics[width=1\linewidth]{figures/shallow_closeup_2.png}
  % \caption{Probability of downstream actions based on different closeup durations.}
  \includegraphics[width=1\linewidth]{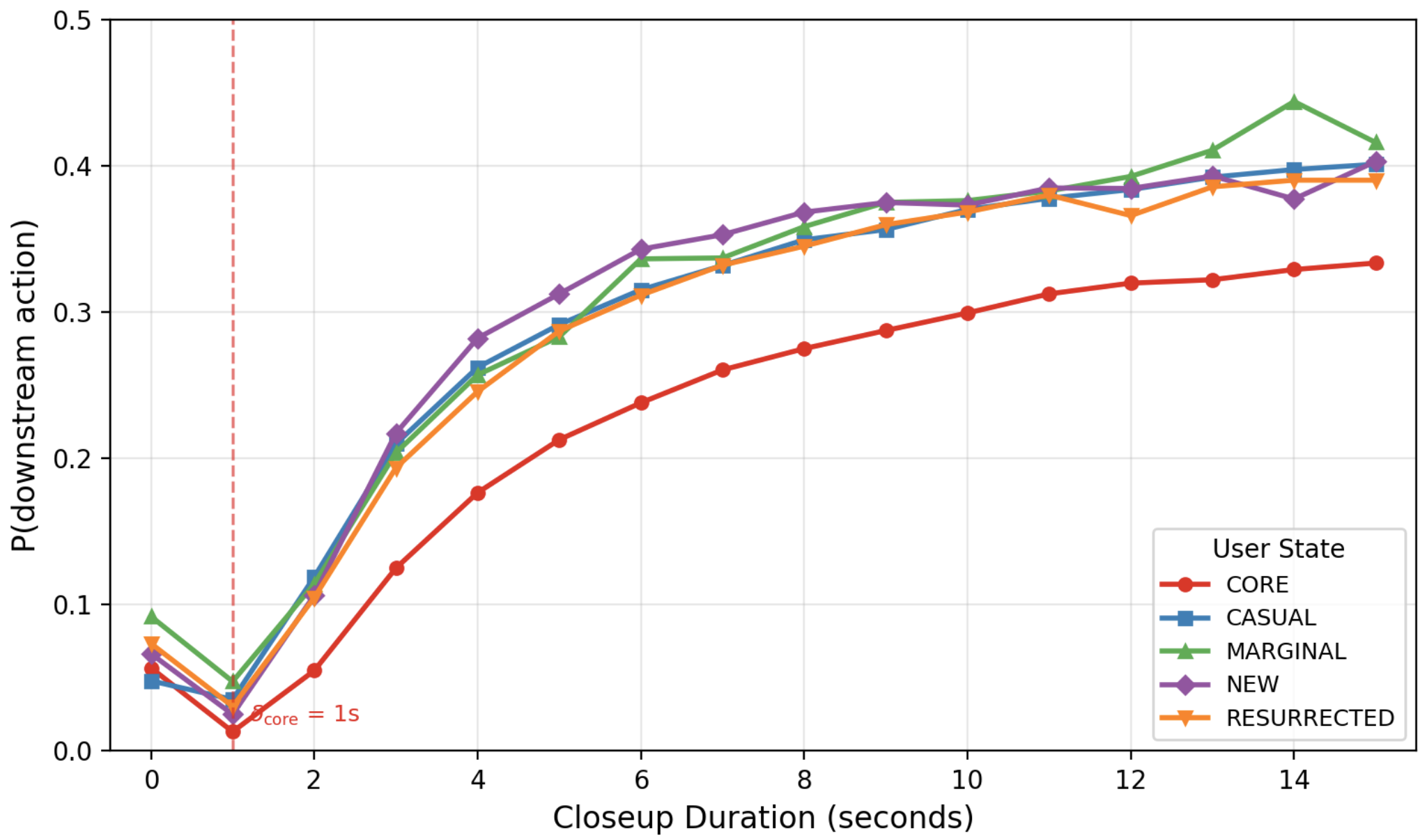}
  \caption{Probability of downstream action (save, clicks, downloads, etc.) as a function of closeup duration based on different user states \footnote{User states group users by their behavioral patterns: for example, users who engage daily and users who engage monthly belong to different user states.}.}
  \label{fig:actionability-by-state}
\end{subfigure}
\caption{Shallow closeups data analysis.}
\label{fig:shallow-closeup}
\end{figure}

While the above downstream rewards focus on positive user actions, we also explicitly model \emph{negative} engagement patterns that are predictive of poor long-term outcomes. An important class of such signals at Pinterest are \emph{shallow closeups}: closeups of a Pin that are quickly abandoned and do not lead to meaningful downstream activity. Empirically, closeups shorter than one second have substantially lower probability of leading to high-intent actions (e.g., saves, long clicks, deeper P2P exploration) and higher probability of immediate session abandonment compared to longer closeups. This is shown in Figure~\ref{fig:shallow-closeup}. At the same time, shallow closeups occur at volumes comparable to major positive signals such as saves, thus indicating high-coverage implicit feedback from user engagements.

For each downstream step $i$ of user $u$, let $d_{u,t+i}$ denote the dwell time if $A_{u,t+i}=\texttt{closeup}$, and let $\mathsf{I}^{\mathrm{pos}}_{u,t+i}\in\{0,1\}$ indicate whether that closeup is followed by any downstream high-intent action within the same trajectory. We define the shallow-closeup indicator
\begin{equation*}
\mathbb{I}^{\mathrm{SC}}_{u,t+i}
=
\mathbb{I}\!\left[
A_{u,t+i}=\texttt{closeup}
\;\wedge\;
d_{u,t+i}<\delta_{\mathrm{state}(u)}
\;\wedge\;
\mathsf{I}^{\mathrm{pos}}_{u,t+i}=0
\right],
\end{equation*}
where $\delta_{\mathrm{state}(u)}$ is a user-state-specific dwell-time threshold.

We choose $\delta_{\mathrm{state}(u)}$ using the offline framework in Section~\ref{sec:insights} to balance label coverage and purity. The actionability curves in Figure~\ref{fig:actionability-by-state} show distinct inflection points for different user segments: for core users, downstream action probability rises sharply around $1$s, while for non-core users the transition occurs closer to $0.5$s. We therefore use $\delta_{\mathrm{core}}=1\mathrm{s}$ and $\delta_{\mathrm{non\text{-}core}}=0.5\mathrm{s}$. As shown in Section~\ref{sec:experiments}, this state-specific design performs better than a uniform threshold.

Negative rewards for a predicted trajectory $\hat{\mathbf{S}}$ can be defined as
\begin{equation*}
    \mathbb{R}^{\text{neg}}(\hat{\mathbf{S}}) = - \lambda^{\text{SC}} \sum_{i=1}^{n_{u,t}} \mathbb{I}^{\text{SC}}_{u,t,i}
\end{equation*}
with $\lambda^{\text{SC}} > 0$ to be the strength of the penalty. The overall surrogate reward in Definition~\ref{def:proxy-p2p} becomes a combination of positive and negative rewards: $\mathbb{R}(\hat{\mathbf{S}}) = \mathbb{R}^{\text{pos}}(\hat{\mathbf{S}}) + \mathbb{R}^{\text{neg}}(\hat{\mathbf{S}})$ where $\mathbb{R}^{\text{pos}}$ represents positive downstream rewards discussed in other sections.
% including $\mathbb{R}^{^{\mathrm{eng}}}$ and $\mathbb{R}^{^{\mathrm{UCA}}}$.

\subsubsection{Rewards for Use Case Adoption}\label{reward-uca}

Beyond rewards for deeper sessions and revisitation, we introduce a downstream reward for use case adoption, which captures whether a recommended item helps a user engage with content outside their established interest space. This reward is motivated by the observation that long-term retention on Pinterest depends not only on reinforcing existing preferences, but also on helping users build connections to new use cases over time.

Specifically, for a user~$u$, we represent historical interests with precomputed interest cluster embeddings $\{c_{u,1},\allowbreak \dots,\allowbreak c_{u,K}\}$ and a candidate item~$x$ with its embedding~$e_x$. We define~$x$ as a new use case if its similarity to every existing cluster is low:
$
\text{new}(u, x) = \mathbb{I}\left[\max_{1 \leq i \leq K} \cos(e_x, c_{u,i}) < \eta\right]
$
where $\eta$ is a tunable threshold and $\mathbb{I}$ is the indicator function. The use case adoption reward is then defined as a conjunctive signal:
$$
\mathbb{R}_{u,x}^{\text{UCA}} =
\begin{cases}
1 & \text{if } \text{engaged}(u, x) \wedge \text{new}(u, x) \\
0 & \text{otherwise}
\end{cases}
$$

A training example is positive only when the item is both engaged and identified as a new use case, and negative otherwise. In this way, the reward encourages the model to surface content that is novel and capable of eliciting meaningful user action, providing a proxy for expanding user intent while remaining aligned with long-term engagement objectives.

\subsection{Rewards Derivation Infrastructure}\label{sec:infra}
Deriving downstream reward (DR) labels from large-scale user interaction logs poses two main challenges: (i) the combinatorial space of reward definitions, and (ii) the need to quickly iterate on new labels without incurring additional engineering effort or training cost. We address these challenges by evolving from a pre-aggregated downstream rewards table (DRv1) with labels generated ahead of training, to a DRv2 infrastructure which stores user engagement as reusable daily sequences and derives labels on demand inside a Ray-based dataloader~\cite{moritz2018ray}.

The initial DRv1 implementation precomputed DR labels and stored them in a table, where each row corresponded to a single impression or closeup event for a user and contained pre-aggregated $n$-hop rewards computed by Spark workflows that repeatedly self-joined engagement logs along Pin to Pin (P2P) trajectories. Any change in the reward definition such as attribution window, hop discounting, and action filters required recomputing the full DRv1 table and backfilling all dependent training datasets across Pinterest, extending iteration cycles to weeks for a single new label variant.

To remove this bottleneck, DRv2 represents user behavior as daily engagement sequences and moves reward computation into the dataloader. A unified DRv2 table stores an ordered sequence of engagement events with aligned arrays of per event attributes such as content identifiers, action types, timestamps, surface identifiers, etc. These sequences are produced once by Pinterest's pre-existing user sequence platform and reused across teams and surfaces. When constructing a training dataset for a surface, we join that surface's recommendation log with the DRv2 table on $(u,\mathrm{dt})$, so that each training example is paired with the subsequent engagement trajectory of the same user on that day across all surfaces. Inside the Ray dataloader, we then compute DR labels on the fly by applying a configurable user-defined function (UDF) to each batch: for every training example, the UDF scans its attached engagement sequence and derives the requested labels.

\begin{figure}[ht]
    \centering
    \includegraphics[width=0.99\linewidth]{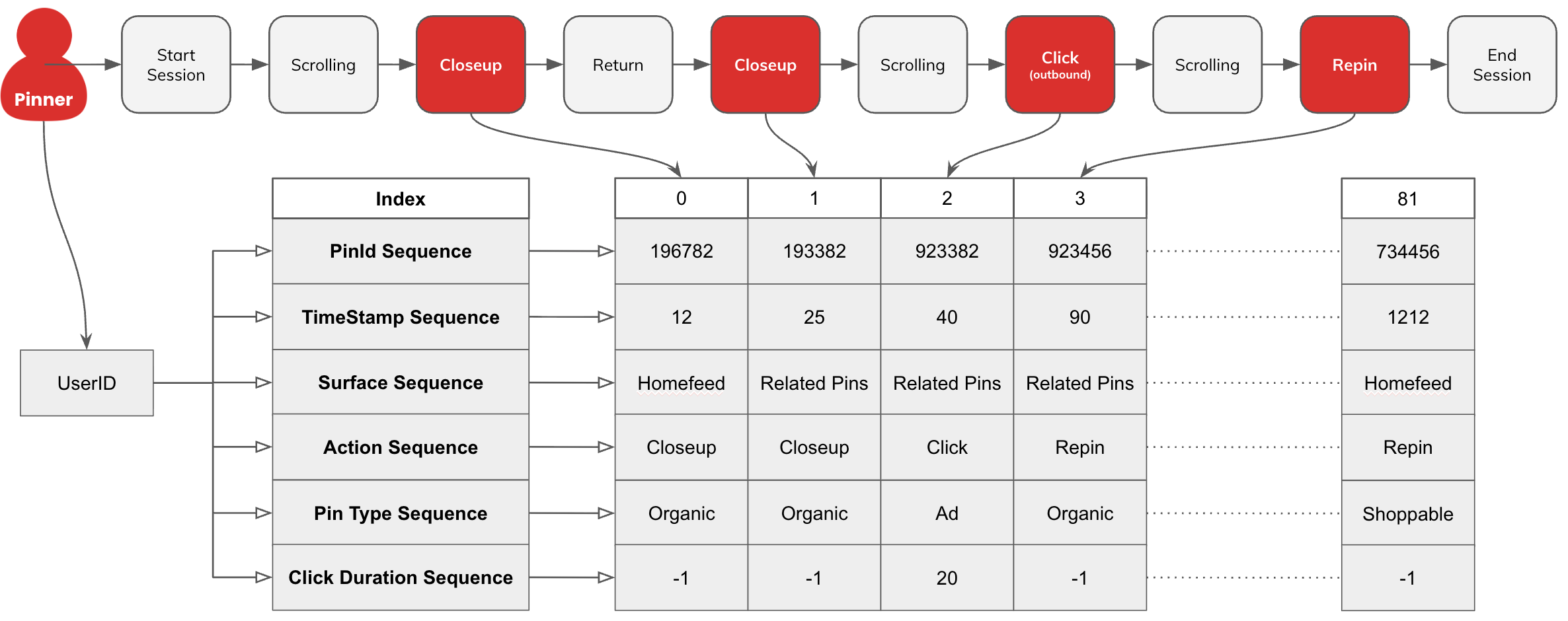}
    \caption{DRv2 table structure illustration: per-user daily event sequences with aligned attribute arrays}
    \label{fig:drv2_infra}
    \Description[DRv2 Infra]{DRv2 infrastructure that generates different label sequences for downstream rewards modeling.}
\end{figure}

Overall, this Ray-based framework optimizes DR label derivation by replacing the Spark self-joins and table recomputation in DRv1 with on-demand computation in the dataloader, while defining reward variants over shared DRv2 engagement sequences. This yields a roughly $10\times$ reduction in idea-to-experiment time from $\sim$3 weeks to $\sim$2 days, keeps incremental training cost within $5\%$ of baseline step time since UDF execution overlaps with data I/O, and makes it practical to rapidly explore complex reward objectives, including multi-day cross-surface and revisitation-based signals.

\subsection{Balance of Immediate and Downstream Rewards}\label{sec:balance}
Our objective is to improve long-term outcomes without degrading immediate user experience. When a downstream reward is implemented as an auxiliary to a multi-head recommendation model, we score item $x$ by linearly combining immediate-engagement and downstream-reward predictions:
$s(x) = \sum_{i=1}^{N} w_i p_i(x),$
where $p_i(x)$ is the prediction of head $i$ and $w_i$ is its serving weight. Unlike DT4IER~\cite{liu2024sequential}, which learns these weights end-to-end, we tune $\{w_i\}$ outside the ranker using HyperOPT~\cite{bergstra2013making,bergstra2015hyperopt}. This decouples weight search from model training, making it easier to add new reward heads and enforce product constraints. In practice, we choose weights to keep immediate engagement neutral while improving retention related metrics. For negative reward heads (Section~\ref{sec:negative-rewards}), the corresponding weight is negative, so a higher predicted probability lowers the ranking score. Other downstream rewards can also be integrated through training-time reweighting.

%% file: sections/6_experiment.tex
\section{Experiments}\label{sec:experiments}
We use online A/B experiments on real-world Pinterest engagement data to answer the following research questions:
\begin{itemize}
    \item \textbf{RQ1:} How does our proposed downstream rewards learning framework help improve short-term and long-term user engagements and retentions under industry settings?
    \item \textbf{RQ2:} How much improvement does each downstream rewards design bring respectively?
    \item \textbf{RQ3:} How effective are our proposed downstream rewards under different surfaces at Pinterest?
\end{itemize}

\subsection{Experiment Setups}
The datasets we use for training and offline evaluation are similar to the ones from \cite{xia2023transact}. All the experiment results are based on Pinterest data, and we do not evaluate our Downstream Rewards methods on public data given the lack of related action signals required for deriving all the kinds of rewards proposed by this paper.
% \kelly{qq: do we want to say this loud?}
Besides, the evaluation of public dataset is limited given they do not necessarily reflect the actual online engagement patterns. The Homefeed online experiments receive 1.5\% of total traffic for 3–4 weeks.

To measure the success of our proposed Downstream rewards modeling frameworks, here are the metrics we use for evaluation:
\begin{itemize}
    \item \textbf{Successful Sessions (SS)}: Number of user sessions on Pinterest with one or more of the following actions: save, search, closeup, create, download, click out etc. This measures the volume of meaningful user engagement journey when they explore Pinterest app or website.
    \item \textbf{Total Time Spent}: Total time spent in app or across different surfaces at Pinterest.
    \item \textbf{DAU, WAU and MAU}: Daily, weekly and monthly active users, which are common metrics for evaluating recommendation quality as well as user retentions.
    \item \textbf{Immediate Action Rates}: Improvement of engagement rates for important actions such as save, closeups, downloads etc. measure the direct user activities.
\end{itemize}
We view these metrics as outcomes that are directionally aligned with the downstream reward signals introduced in Section~\ref{sec:preliminary}. For example, an increase in \textit{Successful Sessions} is consistent with a larger number of user trajectories $\mathbf{S}_{u,t}$ that contain meaningful downstream engagements. Similarly, higher \textit{Total Time Spent} indicate deeper engagement trajectories, which can provide a useful signal for longer-horizon activity.

In addition, these metrics help us assess whether improvements in downstream reward optimization transfer to broader retention impacts. In internal analyses, \textit{Successful Sessions} show positive correlation with retention-related metrics such as WAU and active days in the last 14 days, with Pearson correlation tests having $p < 0.01$. Moreover, improvements in \textit{Successful Sessions}, \textit{Total Time Spent}, and immediate action rates can be interpreted as supportive evidence that the proposed framework improves user engagement quality, while DAU/WAU/MAU provide a complementary view of the gains are reflected in longer-horizon user activity. All the metrics shown below are statistically significant in online experiments.

% \derek{How do these metrics reflect long-term user gains / retentions. Might need to refer to Armando's analysis}

% \derek{We might need to add more details regarding experiment setups.}

\subsection{Experiment Results on Homefeed}
% \derek{Might need a table for all the experiment results, better to have aggregation of results}

\setlength{\abovecaptionskip}{1pt}
\begin{table*}[ht!]
\begin{tabular}{c | c c | c c}
     \specialrule{1pt}{1pt}{3pt}
     & \multicolumn{2}{c|}{\Large \textbf{Successful Sessions}} & \multicolumn{2}{c}{\Large \textbf{Total Time Spent}}\\
     \Large{\textbf{Downstream Rewards}} & Core Users & Non-core Users & Core Users & Non-core Users \\
     \specialrule{1pt}{1pt}{3pt}
    Rewards for deeper sessions & \textbf{+0.24\%} & \textbf{+0.48\%} & \textbf{+0.09\%} & \textbf{+0.10\%} \\
     \midrule
    Negative rewards & \textbf{+0.16\%} & \textbf{+0.16\%} & \textbf{+0.46\%} & \textbf{+0.24\%} \\
     \midrule
    Rewards for use case adoption & \ \textbf{+0.10\%} & \textbf{+0.10\%} & \textbf{+0.16\%} & \textbf{+0.11\%} \\ 
     \bottomrule
\end{tabular}
\caption{Experiment results on Homefeed based on different user states with statistically significant relative lifts.}
\label{table:main-result-homefeed}
\end{table*}

Table~\ref{table:main-result-homefeed} shows results of different downstream rewards over long-term user retention and engagement gains at Homefeed of Pinterest.

\subsubsection{Experiment for Deeper Session Engagement Rewards}
% \derek{Should we add results for DR downloads and DR save respectively? This will make tables to have more contents}
% \derek{Also, thinking of adding experiment results from 1-hop downstream action to n-hop downstream actions, we demonstrate the session depth and longer DR trajectory helps.}
In the online A/B test for deeper session rewards, the experiment groups for adding the rewards based on downloads and screenshots action showed gains on metrics aligned with long-term engagement and retention. Site-wide SS increase by $+0.36\%$ across all users. We also observe improvements in depth and duration of engagement: total time spent in app increases by $+0.10\%$ and overall screenshots by $+0.7\%$ and overall downloads by $+0.7\%$. For deeper session rewards on save actions, we also observed $+0.16\%$ SS gains in online experiments which indicates the effectiveness of this downstream reward on different action signals. WAU increased positively by $+0.1\%$ indicating that rewarding recommendations that lead to deeper session engagement also introduce positive user retention.

\subsubsection{Experiment for Negative Rewards}

We deploy the shallow closeup negative reward on Homefeed as an additional head in the Pinnability~\cite{xia2023transact} ranking model with a negative utility weight. We first use the offline framework to analyze seven candidate treatments formed by three duration thresholds ($<0.5$s, $<1$s, $<2$s) and two label definitions (duration-only vs. duration with no downstream action). The offline analysis results demonstrate that very short thresholds produce too few labels for reliable online impact, while long thresholds introduce noise and weaken revisitation prediction. Based on this analysis, we launch an initial online experiment with a uniform threshold ($\delta=1$s). Although aggregate Homefeed metrics improve, extended measurement shows negative retention and engagement trends for non-core users due to heterogeneity of different user activities. Additional cross surface analysis shows that the treatment increases transitions from Homefeed to P2P, but also reduces engagement from other surfaces, indicating cross-surface cannibalization and a mismatch between a single threshold and heterogeneous user behavior.

% \begin{figure}[h]
%     \centering
%     \includegraphics[width=0.99\linewidth]{figures/shallow_closeup_session_gains.png}
%     \caption{The session > 5min gains for shallow closeups online A/B experiments.}
%     \label{fig:shallow_closeup_sessions}
%     \Description[Session length increase for shallow closeups]{Sessions > 5min increase for 3+ weeks of shallow closeup experiments.}
% \end{figure}

We then conduct an offline analysis by user state and find that the effective threshold differs across segments. This motivates a second launch with thresholds ($\delta_{\mathrm{core}}=1$s, $\delta_{\mathrm{non\mbox{-}core}}=0.5$s) optimized for different user states, followed by utility weight tuning using the HyperOPT. In a month-long experiment, this configuration improves both engagement and retention-related outcomes: SS increase by $+0.16\%$, unsuccessful sessions decrease by $-0.40\%$, total time spent increases by $+0.35\%$, and time spent on Homefeed and Related Pins increases by $+0.2\%$ and $+0.5\%$, respectively. Longer sessions also increase, while hide and report rates decrease. Unlike the initial launch, DAU/WAU trends remain positive for both core and non-core users. These results show that negative rewards are effective to increase session depth of different users. 
% A related metric plot is shown in Figure~\ref{fig:shallow_closeup_sessions}.

\subsubsection{Experiment for Use Case Adoption}
We evaluate the use case adoption reward via a live A/B experiment on homefeed. The reward is implemented by upweighting training examples where the engaged item qualifies as a new use case. Concretely, the positive sample weight is doubled for Pins that are both engaged and satisfy $\mathbb{R}_{u,x}^{\text{UCA}} = 1$, applied across top impactful heads. The cosine similarity threshold $\tau$ is tuned across $\{0.5, 0.6, 0.7\}$, with $\tau = 0.6$ yielding the best balance between label volume and downstream performance.

We observe consistent gains across both engagement and use case adoption metrics. Homefeed saves improve by $+0.42\%$ and Pin clicks by $+0.32\%$, while SS increase by $+0.1\%$ and time spent by $+0.15\%$. Notably, use case adoption propensity on Homefeed improves by $+0.18\%$, and the proportion of weekly active users engaging with two or more use cases increases by $+0.18\%$, directly validating that the reward encourages users to explore content beyond their established interest space. We also observe improvements in deeper actions such as Pin downloads ($+1.28\%$) and screenshots ($+1.02\%$), suggesting that surfacing new use case content drives not only broader exploration but also deeper intent signals.

\subsection{Scaling to Ranking Models in Other
Surfaces}
A user’s journey on Pinterest is inherently non-linear. Users may start on Homefeed, refine intent through Search, continue exploration via Related Pins from a closeup, and later engage with Notifications in subsequent sessions, often traversing multiple surfaces within a single session. This behavior motivates extending downstream reward optimization beyond the entry surface, so that ranking models across surfaces are jointly optimized for the full cross-surface user trajectory. After launching downstream rewards in the Homefeed ranking model, we extended the same framework to Search, Related Pins, and Notifications. In experiments on these surfaces, control models optimize only immediate engagement predictions. Test models optimize both immediate engagement and downstream rewards that capture longer-horizon utility. All experiments receive traffic volume similar to the experiments on Homefeed. We evaluate deeper-session engagement signals as downstream rewards for these ranking models.

\setlength{\abovecaptionskip}{1pt}
\begin{table}[ht]
\centering
\begin{tabular}{l | l | c}
     \specialrule{1pt}{1pt}{3pt}
     \textbf{Surface} & \textbf{Metric} & \textbf{Lift} \\
     \specialrule{1pt}{1pt}{3pt}
     Search          & Search Fulfillment Rate     & \textbf{+0.25\%} \\
     \midrule
     Related Pins    & Successful Sessions           & \textbf{+0.15\%} \\
     \midrule
     Notifications  & Successful Sessions         & \textbf{+0.14\%} \\
     \bottomrule
\end{tabular}
\caption{Online A/B results from extending deeper-session engagement downstream rewards to the ranking models on additional surfaces. Each row compares a test group whose ranker includes downstream reward heads against a control group whose ranker only predicts immediate engagement.}
\label{table:main-result-other-surface}
\end{table}

Table~\ref{table:main-result-other-surface} summarizes the results. On Related Pins, adding downstream reward heads to the P2P ranking model leads to a $+0.15\%$ increase in session frequency, indicating that users return to the app more often when the P2P ranker is optimized for cross surface, long-horizon value rather than only the next click. Besides we have also observed WAU gains after the launch. On Search, the same approach produced a $+0.25\%$ increase in \textit{search fulfillment rate} \cite{agarwal2024omnisearchsage}, defined  as the proportion of searches that result in a user
engagement action of significance (e.g., a click, save etc). For Notifications, in addition for SS gains, we obtain about +0.11\% WAU gains after incorporating the downstream rewards signal. This shows optimizing each ranker on different surfaces for downstream rewards captures complementary parts of the same non-linear user trajectory.

In addition, to verify that the negative reward generalizes beyond the entry surface, we deploy a separate shallow-closeup penalty on the P2P ranker in addition to Homefeed. Because P2P closeup durations are characteristically longer than Homefeed closeups, the offline actionability analysis yields a higher optimal threshold of $\delta_{\mathrm{P2P}} = 2$s (duration-only, without the no positive action condition similar to Homefeed). In a month-long A/B test, the P2P deployment produces $+0.14\%$ SS, and $+0.70\%$ sessions longer than $5$ minutes. 

These results show that downstream rewards can be adopted across distinct ranking models and applications with minimal re-engineering, and doing so optimizes the users journey holistically rather than greedily optimizing each surface in isolation. The simplicity of adoption, combined with consistent gains on both engagement (session frequency, SS) and surface specific quality metrics (search fulfillment rate), makes downstream rewards a practical lever for industrial recommender systems composed of multiple interacting ranking models.
% \subsection{Qualitative Analysis: How Downstream Rewards Change Recommendations}

% To understand the qualitative impact of downstream rewards on the recommendation output, we compare the characteristics of recommended pins before and after incorporating DR signals. Pins with the highest downstream reward scores tend to belong to verticals with strong save-to-retention associations (e.g., Entertainment, DIY \& Crafts, Women's Fashion) and are more likely to elicit deep engagement actions such as saves, long clicks, and P2P exploration. Conversely, pins with the lowest DR scores are disproportionately associated with shallow closeups and single-action sessions.

% After deploying DR, the recommendation system exhibits several observable shifts: (i)~increased diversity of topics within sessions driven by deep engagement rather than shallow browsing, (ii)~a higher proportion of pins that lead to P2P rabbit-hole exploration, and (iii)~a modest increase in the surfacing of fresh pins, likely because newer content that elicits strong engagement receives higher DR-aligned scores. These patterns are consistent across both core and non-core user segments, reinforcing that the proposed downstream rewards generalize beyond a single user population.

%% file: sections/2_related.tex
\section{Related Works}\label{sec:related}
% \filip{Why is this section 6 and not section 2?}
% \derek{We usually want to see main content of the paper starts at Page 2-3, if the related works section is too long, we could potentially make the paper tail heavy in terms of the content to cover.}
\subsection{General Recommendation Systems}
Traditional machine learning algorithms such as gradient boosting \cite{cheng2014gradient,chen2016xgboost,ke2017lightgbm} are strong tabular based methods for recommendations. However, they usually suffer from heavy feature engineering requirements, and they cannot process unstructured modalities such as text and images efficiently. Besides, it is hard to easily adapt these methods for modeling long sequences which are common in modern recommendation tasks.

Deep recommendation models, including Wide \& Deep \cite{cheng2016wide}, DeepFM \cite{guo2017deepfm}, DCN \cite{wang2017deep}, and DCNv2 \cite{wang2021dcn}, improved large-scale ranking by learning feature interactions directly from data. Multi-task objectives further learn multiple user action signals jointly \cite{tang2020progressive,bai2022contrastive,gong2022real,liu2023deep,liu2023multi,zhang2024m3oe}. However, these methods are still primarily optimized for short-horizon engagement targets, such as clicks and other immediate actions, rather than long-term user retention.

% In the past decade, deep learning has been proved to effective solving different tasks. Deep learning based recommendation systems such as Wide \& Deep \cite{cheng2016wide}, DeepFM \cite{guo2017deepfm}, DCN \cite{wang2017deep} and DCNv2 \cite{wang2021dcn} are deployed in different industrial settings and show good performance on real-world large scale recommendation tasks. These models apply different techniques such as feature crossing and feature interaction learning but they originally were designed for user engagement signal such as click-through rate (i.e. CTR). Later, multi-task learning recommendation models \cite{tang2020progressive,bai2022contrastive,gong2022real,liu2023deep,liu2023multi,zhang2024m3oe} are proposed which enable the recommendation systems to capture multiple user action signal at the same time. However, unlike other methods we discuss in the subsection~\ref{subsec:retentino-opt} below, these models mainly optimize for immediate user engagement signals which are not directly related to long-term retention signals. Therefore, it is hard for them to maximize user retention which is crucial for modern recommendation systems and could potentially result in increase of Daily Active Users (i.e. DAU) and revenue gains.

In recent years, more complex recommendation paradigm \cite{linkedin,evnine2024achieving} became popular in different industries. This type of multi-stage recommendation system consists of different components including retrieval, ranking, re-ranking and blending. Each components have different responsibility and advantages. For instance, retrieval models are usually efficient and can filter billions of items to thousands of top relevant candidates, while re-ranking and blending algorithms can help fulfill different business objectives such as adding diversity on topics etc. With the rise of Large Language Models \cite{schulman2022introducing}, large scale generative ranking algorithms \cite{zhai2024actions,jiang2025recgpt,ju2025generative,wang2025scaling,zhu2025rankmixer} have sought to unify multiple components of the ranking pipeline within a single, billion-parameter model trained on web-scale data. While these methods have shown promising online results, these approaches are computational heavy and lack the flexibility required for deployment in different specific use cases such as home page recommendations or related items recommendations. Moreover, optimizing user retention remains challenging in the end-to-end generative ranking framework.

\subsection{User Retention Optimization}\label{subsec:retentino-opt}
Prior work on retention optimization in recommendation systems spans several directions. Contextual bandit and reinforcement learning (RL) methods \cite{wu2017returning,han2019optimizing,pei2019value,zou2019reinforcement,zhang2021counterfactual,wang2022surrogate,zhang2022multi,cai2023reinforcing,xu2023optimizing,xue2023prefrec,xue2025auro} optimize cumulative rewards to identify actions with long-term user value. However, these approaches typically require carefully designed reward functions and high-quality feedback data. Moreover, different applications impose different objectives and system constraints, making it difficult to standardize reward design within a single RL framework. Other works use sequence based modeling to improve user retention \cite{devooght2017long,zhao2020maximizing,zhao2023user,liu2024sequential}. Related approaches also construct item graphs to infer user interest clusters for sequential recommendation \cite{cen2020controllable,tian2022multi,zhang2022re4,du2024disentangled,chunjing2025learning}. 
% While these methods capture temporal dynamics, sequence based approaches still face causal attribution challenges when linking retention outcomes to earlier recommended slates.
In contrast, our model-agnostic downstream reward formulation uses supervised signals from observed trajectories as a practical surrogate for the long-horizon rewards. This avoids REINFORCE-style policy gradient optimization, which is often difficult to train and deploy across surfaces. In addition, Section~\ref{reward-uca} uses pre-extracted user-interest clusters from upstream services, without requiring additional graph construction or joint training for sequence modeling. Similar ideas appear in \cite{chang2023twin,si2024twin}.

Recent work has studied retention through causal and explicit return signal modeling \cite{du2019improve,zhang2021user}. Examples include predicting future return probability \cite{wang2025not}, adding survey based return intent into ranking models \cite{bakhshi2025retentive}, and using revisitation of saved content as an auxiliary retention target \cite{jiang2026save}. Our work is complementary to these approaches: rather than relying on a single proxy, we present a unified framework for discovering, comparing, and combining heterogeneous downstream rewards, motivated by the observation that no single signal fully captures long-term user value on a visual content discovery platform. Closest to our setting, \cite{wang2022surrogate} identifies user behavior patterns associated with long-term experience, but its reward concepts, such as ``high-quality consumption'' and ``repeated consumption'' defined by time spent and revisitation of the same item, do not transfer directly to Pinterest, where content is predominantly visual, user intent is often inspiration driven, and actions like save and download are especially informative. Reward design at Netflix \cite{tang2023reward} also targets long-term satisfaction, but is developed for a bandit-feedback setting with a specific proxy reward. In contrast, our downstream reward framework is more general and can be more easily adapted to other recommendation settings.

%% file: sections/7_conclusions.tex
\section{Conclusions}\label{sec:conclusion}
In this paper, we presented a model-agnostic downstream reward framework for optimizing long-term user engagement and retention in large scale recommendation systems. Guided by offline analysis on real-world Pinterest interaction data, we developed downstream rewards that capture deeper session engagement, negative engagement patterns, and use-case adoption beyond a user’s established interests. We further described an efficient infrastructure for large scale reward-label generation and a practical serving strategy that balances immediate engagement with downstream value. Online A/B experiments show that these rewards consistently improve engagement and retention related metrics, and the framework has been deployed across multiple surfaces at Pinterest. Promising directions for future work include more sophisticated session-level reward design such as revisitation and tighter integration of these rewards into ranking models beyond auxiliary prediction signals.

%% file: sections/8_acknowledgement.tex
\section{Acknowledgements}
% We would like to thank Jaewon Yang, Haibin Xie, Alekhya Pyla, Zhongjian Jiang for their technical feedback and help in adopting the downstream rewards across Pinterest.
We would like to thank Jaewon Yang for his technical feedback on this paper.